\documentclass{article}

\usepackage{PRIMEarxiv}

\usepackage[utf8]{inputenc} 
\usepackage[T1]{fontenc}    
\usepackage{hyperref}       
\usepackage{url}            
\usepackage{booktabs}       
\usepackage{amsfonts}       
\usepackage{nicefrac}       
\usepackage{microtype}      
\usepackage{lipsum}
\usepackage{graphicx}
\graphicspath{{media/}}     
\newcommand{\fa}{FA}
\newcommand{\cb}{CB}
\newcommand{\HTTP}{HTTP}
\usepackage{cleveref}
\usepackage{float}
\usepackage{xcolor}
\definecolor{loudorange}{RGB}{255,165,0}
\definecolor{silentred}{RGB}{178, 34, 34}
\newcommand{\firsthand}{\textsuperscript{*}}
\usepackage{listings}
\usepackage{xcolor}

\title{FailureAtlas: A Taxonomy of Failure Modes in Multi-Provider LLM Serving Infrastructure}

\author{
  Vishal Pandey\\
  Metriqual \\
  London, UK \\
  \texttt{vishal@metriqual.com} \\
   \And
  Gopal Singh\\
  Metriqual \\
  Athens, GR \\
  \texttt{gopal@metriqual.com} \\
}

\begin{document}
\maketitle

\begin{abstract}
Multi-provider LLM gateways reverse proxies that route, load-balance, and rate-limit requests across foundation-model APIs have become critical production infrastructure.  Yet the failure modes specific to this architectural layer remain undocumented, scattered across issue trackers and post-mortems with no unifying framework.  We introduce \fa{}, a two-axis taxonomy that classifies failures by their \emph{origin layer} (Network/Transport, Streaming/Protocol, State/Session, Model~Behavior, Governance/Cost) and their \emph{detectability} (Loud vs.\ Silent).  We populate this taxonomy with five verified catalog entries sourced from public bug reports and first-hand stress testing, each accompanied by a mechanistic root-cause analysis. Three entries include standalone reproduction scripts. Our principal finding is that the most operationally severe failures are \emph{silent}: they return HTTP~200, pass every standard health check, and corrupt application state in ways that require semantic-level observability to detect. Two such silent failures a concurrency race condition causing history loss and a streaming index collision corrupting tool-call payloads were discovered first-hand during \cb{} evaluation campaigns.
\end{abstract}

\section{Introduction}
\label{sec:intro}
During the development of \cb{}~\cite{pandey2026continuitybenchbenchmarksystemsstudy}, a benchmark for measuring multi-turn continuity in LLM-powered agents, we ran a sustained stress-testing campaign against a multi-provider gateway serving hundreds of concurrent evaluation sessions. Two failures emerged that we did not anticipate, could not find documented anywhere, and most critically could not detect with any standard monitoring tool. The first was a concurrency race condition in conversation-state management. Two coroutines servicing the same session each read the conversation history, appended a new turn, called the model, and wrote the history back.  The second writer silently overwrote the first's contribution.  The result: conversation turns vanished from the context window, the model's generation quality degraded, and the system reported HTTP~200 on every request.  No metric latency, error rate, pod health registered any anomaly.  We discovered the bug only because \cb{}'s \emph{semantic} continuity metrics showed unexplained drops in persona adherence during high-concurrency runs. 

The second was a retry storm.  When the upstream provider returned a transient 502, all 100 parallel evaluation agents retried on the same fixed interval. They synchronised into a thundering herd that saturated the provider's rate limit on every retry window, causing 25\% of requests to fail permanently despite the underlying issue being a momentary hiccup. We independently rediscovered this classic distributed-systems problem~\cite{corbet2003thundering} in the specific setting of LLM gateway clients before later finding fragments of the same pattern scattered across unrelated GitHub issues with no cross-referencing or shared vocabulary. 

These two experiences crystallised an observation: \textbf{multi-provider LLM gateways are now common production infrastructure, but their failure modes are undocumented, unsystematised, and independently rediscovered by each team that encounters them.} 

The infrastructure class itself is recent.  Projects like LiteLLM\footnote{\url{https://github.com/BerriAI/litellm}}, Portkey\footnote{\url{https://github.com/Portkey-AI/gateway}}, and OpenRouter emerged in 2023--2024 to solve a real problem: abstracting over the proliferating landscape of foundation-model APIs with unified routing, load balancing, failover, rate limiting, and cost governance.  They occupy a novel architectural position a reverse-proxy layer between application code and model providers that inherits failure modes from both distributed systems (network partitions, semaphore leaks, thundering herds) and LLM-specific concerns (streaming token protocols, tool-call index tracking, conversation-state coherence). 

Yet no systematic framework exists for this specific failure surface. The distributed-systems community has decades of taxonomic infrastructure: Byzantine fault models~\cite{lamport1982byzantine}, the CAP theorem~\cite{brewer2000cap}, empirical studies of thousands of cloud-system bugs~\cite{gunawi2014bugs}. The LLM evaluation community has HELM~\cite{helm2023}, MMLU, and domain-specific benchmarks. Neither body of work addresses the \emph{intersection} the failures that emerge specifically from serving LLMs through multi-provider gateway infrastructure.  Practitioners encountering a novel failure in this layer must re-derive the root cause from scratch, often misled by surface symptoms.  A \texttt{JSONDecodeError} gets attributed to ``network issues'' when the true cause is a streaming-protocol violation three layers up the stack.  A Redis counter that never decrements gets debugged as a configuration error rather than recognised as a resource-leak pattern triggered only under upstream provider failures.

\fa{} addresses this gap.  We contribute:

\begin{enumerate}
    \item A \textbf{two-axis taxonomy} (Layer $\times$ Detectability) that provides a structured vocabulary for classifying LLM infrastructure failures.  The five layers (Network/Trans\-port, Streaming/Pro\-tocol, State/Ses\-sion, Model~Behavior, Governance/Cost) span the full stack of a multi-provider serving pipeline; the two detectability classes (Loud vs.\ Silent) encode the critical operational distinction between failures that trigger alarms and failures that pass every health check while corrupting the semantic payload~(\Cref{sec:taxonomy}).
    \item An \textbf{evidence-grounded catalog} of five concrete failure modes, each with a mechanistic root-cause analysis and explicit evidence provenance.  Two entries originate from first-hand stress testing; three are sourced from verified public bug reports.  Three of the five include standalone reproduction scripts~(\Cref{sec:case-studies,sec:catalog}).
    \item \textbf{Two fully reproduced case studies} the concurrency race condition and the retry storm characterised in sufficient depth for independent replication: mechanism, minimal reproduction code, verified output, and mitigation~(\Cref{sec:case-studies}).
    \item An \textbf{open, extensible catalog repository}%
          \footnote{\url{https://github.com/Vishal-sys-code/failure-atlas}}
          with a structured YAML schema, automated rendering pipeline, and
          contribution guidelines, designed for community extension as practitioners encounter and document new failure modes.
\end{enumerate}

\noindent
Our principal finding is that the most operationally severe failures in this infrastructure class are \emph{silent}.  They return HTTP~200, pass every standard health check, and corrupt application state in ways that require semantic-level observability continuous evaluation of output quality, conversation-state integrity, and tool-call payload correctness to detect. The industry's current monitoring practices, built around latency percentiles, error rates, and uptime SLAs, are structurally blind to this failure class.

\section{Related Work}
\label{sec:related}

\fa{} sits at the intersection of three bodies of prior work.  We draw on all three but address a gap that none of them covers: the specific failure surface created by multi-provider LLM gateway infrastructure.
\subsection{Distributed-Systems Failure Taxonomies}
\label{sec:related-ds}
The systematic classification of failures in distributed systems is mature. Lamport, Shostak, and Pease~\cite{lamport1982byzantine} established the Byzantine failure model, distinguishing crash faults from arbitrary (potentially adversarial) deviations. Brewer's CAP theorem~\cite{brewer2000cap} formalised the fundamental trade-off between consistency and availability under network partitions.  Gunawi et~al.~\cite{gunawi2014bugs} conducted a large-scale empirical study of over 3,000 bugs in cloud infrastructure (Cassandra, HDFS, MapReduce), finding that the majority of catastrophic failures were caused by incorrect error handling a pattern we observe directly in the Redis semaphore leak described in \Cref{sec:redis}, where a missing \texttt{finally} block on an error path causes a permanent rate-limit deadlock.

\fa{} inherits the \emph{taxonomic method} from this tradition classify by fault origin and observability but targets a system layer that did not exist when these frameworks were developed. Classical distributed-systems taxonomies model failures in storage, consensus, and message-passing systems. They do not account for the concerns specific to LLM serving: stateful conversation histories as a critical data structure, streaming token protocols with inter-chunk index dependencies, and the fundamental asymmetry between ``infrastructure correctness'' (HTTP~200, healthy pod) and ``semantic correctness'' (the model received the right context and produced a coherent response).  The Detectability axis in \fa{} Loud vs.\ Silent is our attempt to encode this asymmetry into the taxonomy itself: a dimension that would be unnecessary in a system where infrastructure-level success reliably implies application-level success, but is essential in one where it does not.

Several of the failure modes we catalog are recognisable as classical distributed-systems problems wearing new clothes.  The retry storm (\Cref{sec:retry-storm}) is a thundering herd~\cite{corbet2003thundering}; the Redis semaphore leak (\Cref{sec:redis}) is a resource-leak liveness failure; the race condition (\Cref{sec:race-condition}) is a read-modify-write anomaly. The contribution is not in identifying these patterns as novel in the abstract, but in documenting their concrete manifestation in the specific context of LLM gateway infrastructure, where their consequences (silent context corruption, runaway token spend, permanent API-key lockout) are domain-specific and their detection requires domain-specific tooling.
\subsection{LLM Evaluation and Reliability}
\label{sec:related-eval}

The LLM evaluation literature is extensive but narrowly scoped relative to the concerns of \fa{}. Benchmarks such as HELM~\cite{helm2023}, MMLU, HumanEval, and domain-specific suites measure \emph{model-level} capabilities: factual accuracy, reasoning, code generation, toxicity, fairness. They treat the model as a stateless function a black box that maps a prompt to a completion and evaluate single-turn or few-turn outputs in isolation. Infrastructure is assumed to be transparent; the implicit contract is that if you send a prompt, you receive the model's output, unmodified, with nothing lost or corrupted in transit.

\fa{} documents the cases where this contract breaks. The streaming index collision (\Cref{sec:ollama}) delivers the model's output to the client, but with a corrupted \texttt{index} field that causes independent tool calls to be merged into malformed JSON.  The race condition (\Cref{sec:race-condition}) delivers the model's output correctly, but against an incomplete context from which prior turns have been silently dropped. In both cases, the model itself performed correctly; the failure is entirely infrastructural.  No model-level benchmark would detect either bug, because model-level benchmarks evaluate the model's \emph{capability}, not the \emph{fidelity} of the infrastructure delivering that capability to the application.

More recent evaluation work has begun to move beyond single-turn capability measurement.  Long-context benchmarks (RULER, Needle-in-a-Haystack) test whether models can retrieve information from extended inputs, but still treat the serving infrastructure as transparent. Agent evaluation frameworks (SWE-bench, WebArena) test end-to-end task completion in multi-step settings, but attribute all failures to the agent or model rather than distinguishing infrastructural from cognitive failure modes. \fa{} occupies the gap between these approaches: it is concerned specifically with the failures that \emph{infrastructure} introduces between a correctly functioning model and a correctly written application.
\subsection{ContinuityBench}
\label{sec:related-cb}
\cb{}~\cite{pandey2026continuitybenchbenchmarksystemsstudy} is a benchmark for measuring multi-turn \emph{continuity} the degree to which an LLM maintains coherent persona, factual recall, and behavioural consistency across extended dialogue sequences.  \fa{} and \cb{} are companion projects with a deliberate division of labour, and it is worth stating the relationship precisely to avoid confusion.

\cb{} measures one specific failure mode \emph{deeply}: continuity loss in multi-turn conversations.  It provides a quantitative metric (the continuity score), a structured evaluation protocol, and a benchmark dataset.  It answers the question: ``How much state coherence does this model (or system) lose over $N$ turns?''  It does not attempt to classify \emph{why} the continuity was lost whether the cause was model drift, context truncation, a race condition in state management, or a streaming protocol bug that corrupted the conversation history before it reached the model.

\fa{} operates at lower depth but broader scope.  It catalogs \emph{many} failure modes across the full serving stack, each characterised with a mechanistic root-cause analysis but not measured with the sustained quantitative rigour that \cb{} applies to continuity specifically.  It answers a different question: ``What are the categories of things that can go wrong between the application and the model, and which of them are invisible to standard monitoring?''

The two projects are connected by more than authorship.  Two of the five \fa{} catalog entries the concurrency race condition and the retry storm were \emph{discovered} during \cb{} evaluation campaigns. In both cases, \cb{}'s semantic continuity metrics served as the detection mechanism that standard infrastructure monitoring missed. This is not incidental; it is the core argument of both projects. \cb{} demonstrates \emph{that} silent failures exist and can be measured; \fa{} catalogs \emph{what} those failures are and where they originate in the stack. Together, they make the case that the LLM serving layer requires both a taxonomy of failure modes \emph{and} continuous semantic observability to catch the ones that infrastructure metrics cannot.
\subsection{LLM Gateway Software}
\label{sec:related-gateways}
Projects such as LiteLLM, Portkey, and OpenRouter provide unified APIs across multiple LLM providers, offering load balancing, failover, rate limiting, streaming translation, and cost tracking. These projects are production-grade infrastructure with significant adoption LiteLLM alone has over 18,000 GitHub stars and is deployed by organisations routing millions of API calls per day.

Despite this adoption, no systematic study has characterised the failure modes that emerge from using these gateways in production.  The projects maintain public issue trackers, and individual bugs are reported and fixed in the normal course of open-source development.  But each bug is treated as an isolated incident: filed, patched, closed.  There is no shared vocabulary across projects (a ``semaphore leak'' in LiteLLM is independently rediscovered as a ``counter drift'' in a proprietary gateway), no cross-referencing between related failure modes (the retry storm and the semaphore leak are filed under different components despite sharing a common trigger: upstream provider failures), and no classification of which bugs are operationally dangerous because they are \emph{silent}.

\fa{} draws directly from these issue trackers as primary evidence sources, treating verified bug reports as the empirical data of the catalog. Three of our five entries originate from the LiteLLM tracker. This reflects LiteLLM's transparency and market adoption, not a claim that other gateways are immune to similar failures.  We note that the \emph{patterns} synchronous calls blocking async loops, stateless chunk processing losing inter-chunk state, unreleased resource counters are architectural rather than implementation-specific, and are likely to recur in any gateway built on similar design choices.

\section{Taxonomy Design}
\label{sec:taxonomy}
A useful failure taxonomy must satisfy two constraints: it must be \emph{complete} enough that a practitioner encountering a novel bug can locate the correct cell without forcing the classification, and \emph{actionable} enough that the classification itself suggests where to look for the root cause and what kind of tooling is needed to detect it. \fa{} achieves this with two orthogonal axes Layer and Detectability whose intersection produces a $5 \times 2$ matrix (\Cref{fig:grid}).

We considered and rejected several alternative designs before arriving at this structure.  A single-axis taxonomy organised by symptom (``data corruption,'' ``availability loss,'' ``cost overrun'') conflates failures with very different root causes and very different fixes: a retry storm and a Redis semaphore leak both produce 429 errors, but one requires client-side jitter and the other requires server-side resource cleanup. A three-axis taxonomy adding a ``severity'' dimension proved redundant in practice: severity is almost entirely determined by detectability, as we argue below.  The two-axis design is the minimal structure that captures the two questions a practitioner actually asks when triaging a production incident: \emph{where in the stack did this break?} and \emph{why didn't we catch it sooner?}
\subsection{Axis 1: Layer (Origin of Failure)}
\label{sec:layers}
The Layer axis maps the serving stack of a modern LLM-powered application, ordered from lowest to highest abstraction.  We identify five layers. The boundaries are drawn at points where the failure mechanism, the responsible code owner, and the appropriate mitigation strategy all change simultaneously a pragmatic criterion that avoids both over-splitting (which produces empty cells) and under-splitting (which lumps unrelated bugs together).

\begin{description}
    \item[L1 (Network/Transport):] Failures at the lowest level of communication between the application and the model provider: TCP timeouts, TLS handshake failures, DNS resolution issues, and specific to the async-Python ecosystem that dominates LLM gateway implementations synchronous calls blocking asynchronous event loops. These are the closest analogues to classical distributed-systems faults and the most likely to be caught by existing infrastructure monitoring. The event-loop block cataloged in \Cref{sec:healthcheck} falls here. It is architecturally identical to the well-known Node.js event-loop-blocking anti-pattern, transposed to Python's \texttt{asyncio} runtime a reminder that LLM gateways inherit the failure modes of whatever concurrency model they are built on.
    \item[L2 (Streaming/Protocol):] Failures specific to the transport of token streams or chunked responses. This layer exists because LLM APIs have adopted Server-Sent Events (SSE) as their dominant streaming mechanism, introducing a class of stateful protocol bugs that do not arise in traditional request response REST APIs. The critical distinction is that SSE streaming requires both client and proxy to maintain \emph{inter-chunk state}: a running index for parallel tool calls, an accumulator for partial JSON arguments, a notion of ``this chunk continues the same function call as the previous chunk.''  Any proxy layer that processes chunks statelessly treating each SSE event as independent risks corrupting the reassembled payload.  The index collision cataloged in \Cref{sec:ollama} is a direct consequence of this design error.
    \item[L3 (State/Session):] Failures in maintaining context or coherence across a sequence of interactions.  In multi-turn LLM applications, the conversation history is not merely metadata; it is the \emph{primary input} to the model.  Any corruption, truncation, reordering, or silent deletion of turns in this history directly degrades generation quality, because the model is literally operating on incomplete or incorrect data. This layer is uniquely sensitive to concurrency.  A traditional web application serving two concurrent requests to the same user might produce a stale read or a duplicate write, both of which are detectable by standard consistency checks.  An LLM application serving two concurrent requests to the same session can silently drop an entire conversation turn, producing output that is subtly wrong in ways that require \emph{semantic} evaluation to detect the model simply generates a slightly less coherent response, indistinguishable from normal stochastic variation to any metric that does not inspect the history itself.
    \item[L4 (Model Behavior):] Failures intrinsic to the model's generation, reasoning, or adherence to instructions: silent drift in output formatting across provider versions, gradual instruction-following degradation, hallucinations, and unannounced changes to model behaviour during capacity-management events (quantisation, routing to a smaller variant). We include this layer in the taxonomy for completeness despite being unable to populate it with evidence-grade entries (\Cref{sec:empty-cell}). Its inclusion is deliberate: the empty cells in \Cref{fig:grid} are themselves informative, highlighting a category of failure that is widely discussed anecdotally but resistant to the kind of mechanistic, reproducible documentation that the other four layers admit.
    \item[L5 (Governance/Cost):] Failures related to operational boundaries, resource accounting, and financial constraints.  This layer captures the control-plane failures of LLM serving: retry policies that amplify rather than absorb transient errors, rate-limit counters that leak and never recover, cost-tracking mechanisms that fail to account for partial completions or retry-induced duplication.
    
    Both governance entries in our catalog the retry storm(\Cref{sec:retry-storm}) and the Redis semaphore leak (\Cref{sec:redis}) share a common trigger: upstream provider failures.  This is not coincidental.  Governance mechanisms are designed and tested against the happy path; they are most needed, and most likely to fail, precisely when the upstream provider is degraded the worst possible moment for the control plane to malfunction.
\end{description}
\subsection{Axis 2: Detectability}
\label{sec:detectability}

The Detectability axis is the more important of the two.  The Layer axis tells you \emph{where} to look for a bug; the Detectability axis tells
you \emph{whether you will ever look at all}.

\begin{description}
    \item[Loud:] Failures that produce an observable signal in standard production monitoring: HTTP 5xx errors, unhandled exceptions, container crashes, hard rate-limit rejections (HTTP~429), health-check timeouts.  These failures have high signal-to-noise ratios, generate immediate stack traces, and trigger on-call alerts.  They are unpleasant but \emph{tractable}: the incident-response pipeline alert, triage, root-cause, fix, post-mortem is designed for exactly this class of failure and handles it well.
    \item[Silent:] Failures that return HTTP~200~OK and pass every standard health check, yet fail fundamentally at the semantic or task level. They corrupt the payload dropping conversation turns, merging independent tool-call arguments into malformed JSON, or substituting a cheaper model variant in ways that are invisible to any monitoring system that inspects only status codes, latency percentiles, and error rates.
\end{description}

\paragraph{Why detectability determines operational severity:} The practical consequence of this distinction is stark and asymmetric. Loud failures get fixed \emph{because} monitoring catches them. They enter the incident-response pipeline within minutes, attract engineering attention, and are typically resolved within hours or days. The three Loud entries in our catalog the event-loop block (\Cref{sec:healthcheck}), the Redis semaphore leak (\Cref{sec:redis}), and the retry storm (\Cref{sec:retry-storm}) were all reported as production incidents with immediate operational impact, investigated promptly, and patched or mitigated in subsequent releases.

Silent failures persist \emph{because} nothing flags them.  They do not enter the incident-response pipeline.  They do not trigger alerts. They do not attract engineering attention.  They accumulate as invisible technical debt, silently degrading user experience over time.  The two Silent entries in our catalog the race condition (\Cref{sec:race-condition}) and the index collision (\Cref{sec:ollama}) exemplify this pattern:
 
\begin{itemize}
    \item The race condition was discovered only because \cb{}'s semantic continuity metrics showed unexplained drops in persona adherence during high-concurrency stress tests.  In a production deployment without continuous semantic evaluation, this bug would manifest as a vague, intermittent degradation in ``model quality'' that no one could attribute to a specific cause.
    \item The index collision was reported only after a developer manually inspected the raw SSE chunks during debugging of a seemingly unrelated \texttt{JSONDecodeError}.  The failure's symptom malformed JSON on the \emph{next} turn misdirected debugging toward the network layer for an extended period before the streaming-protocol root cause was identified.
\end{itemize}

This creates a survivorship bias in the engineering record. The bugs that get found, reported, and fixed are disproportionately Loud, because Loud is what monitoring is built to detect.  Silent failures are systematically under-represented in issue trackers, post-mortems, and reliability metrics not because they are rare, but because the infrastructure for \emph{noticing} them does not yet exist in most deployments. \fa{}'s Detectability axis is designed to make this bias visible: by forcing every cataloged entry into a Loud or Silent classification, the taxonomy itself surfaces the question that standard monitoring suppresses.

\section{Methodology}
\label{sec:methodology}
The \fa{} catalog is designed to be empirical, not speculative. Every failure mode included in the taxonomy must be grounded in verified evidence rather than theoretical plausibility.  This constraint shapes our evidence-sourcing process, our inclusion criteria, and our approach to building standalone reproductions.
\subsection{Evidence Sourcing}
\label{sec:methodology-sourcing}
We sourced candidate failure modes from two distinct streams: first-hand stress testing and structured issue-tracker surveys.

\paragraph{First-hand stress testing:} Two of the five catalog entries the concurrency race condition (\Cref{sec:race-condition}) and the retry storm (\Cref{sec:retry-storm}) originate from first-hand observation during \cb{}~\cite{pandey2026continuitybenchbenchmarksystemsstudy} Phase~2 evaluation campaigns. These campaigns ran hundreds of parallel agent instances against a mock LLM-provider gateway designed to simulate real-world latency, rate limits, and transient 502 errors. Both failures emerged organically under sustained load and were documented contemporaneously via system logs and semantic-continuity drops. They are marked with \firsthand{} in the taxonomy grid.

\paragraph{Issue-tracker surveys:} The remaining three entries were sourced from a structured survey of the public GitHub issue trackers for three widely adopted multi-provider LLM gateways: LiteLLM, Portkey, and OpenRouter.  We targeted specific layers of the taxonomy using scoped search queries designed to surface concrete infrastructure-level failure modes rather than general usage questions. For example, to populate the Streaming/Protocol layer, we queried for terms such as \texttt{SSE stream "cut off" OR "truncated"} and \texttt{function calling "malformed JSON" streaming}.  To populate the Network/Transport layer, we queried for \texttt{"connection timeout" gateway failover} and \texttt{asyncio block}.

This targeted search yielded dozens of candidate reports.  However, the vast majority were filtered out during the validation phase because they failed to meet our strict inclusion criteria.
\subsection{Inclusion Criteria and Validation}
\label{sec:methodology-validation}

To be accepted into the \fa{} catalog, a candidate failure report had to satisfy three inclusion criteria:
\begin{enumerate} 
    \item \textbf{Concrete and specific:} The report could not be a general complaint (e.g.,\ ``the model got worse'' or ``the connection dropped'').  It had to describe a specific failure mode occurring under identifiable conditions.
    \item \textbf{Independently verifiable provenance:} The failure had to be documented by a persistent, verifiable URL pointing to a primary source (a GitHub issue, a post-mortem, or a merged pull request fixing the bug). Anecdotal claims without supporting logs or code references were rejected.
    \item \textbf{Mechanistically explainable:} The root cause of the failure had to be fully understood and explainable at the system level. A report of a crash with a stack trace was insufficient unless the sequence of events leading to the crash the *mechanism* could be unambiguously determined.
\end{enumerate}

The validation process was manual and rigorous. For each candidate, we read the entire issue thread, examined the provided logs, traced the execution path through the relevant open-source codebase, and evaluated the merged fix (if available) to confirm that the described mechanism accurately matched the code.  The three surveyed entries in our catalog the event-loop block (\Cref{sec:healthcheck}), the index collision (\Cref{sec:ollama}), and the Redis semaphore leak (\Cref{sec:redis}) are the survivors of this filtering process. They represent the gold standard of our inclusion criteria: real bugs,
affecting production users, with proven root causes.

\subsection{Standalone Reproductions}
\label{sec:methodology-reproductions}
Where possible, we aim to provide standalone reproduction scripts that demonstrate the failure mechanism in a minimal, isolated environment. Of our five catalog entries, three include full reproductions in the \fa{} repository.

Building these reproductions requires isolating the core logic from the surrounding infrastructure. For the race condition (\Cref{sec:race-condition}), we built a minimal `asyncio` script that simulates shared vs.\ isolated state under concurrent writes, stripping away the full \cb{} harness to expose the raw read-modify-write anomaly. For the retry storm (\Cref{sec:retry-storm}), we implemented amock rate-limited provider and ran 100 concurrent simulated agents against it, demonstrating the failure of fixed-interval retries side-by-side with the success of exponential backoff with jitter.

For the Ollama index collision (\Cref{sec:ollama}), we adapted the exact reproduction script provided by the original reporter in the LiteLLM issue tracker.  We modified it to run against statically captured chunk data, confirming that the proxy's iterator incorrectly emits \texttt{index=0} and triggers a \texttt{JSONDecodeError} without requiring a live Ollama server.

The remaining two entries the event-loop block and the Redis semaphore leak do not include standalone scripts because their failure mechanisms are inextricably tied to external infrastructure (Kubernetes health probes and a running Redis instance, respectively).  For these, we provide detailed `README.md` instructions detailing the environmental setup required to reproduce the failure.

\section{The Catalog}
\label{sec:catalog}
\begin{figure}[t]
    \centering
    \includegraphics[width=\textwidth]{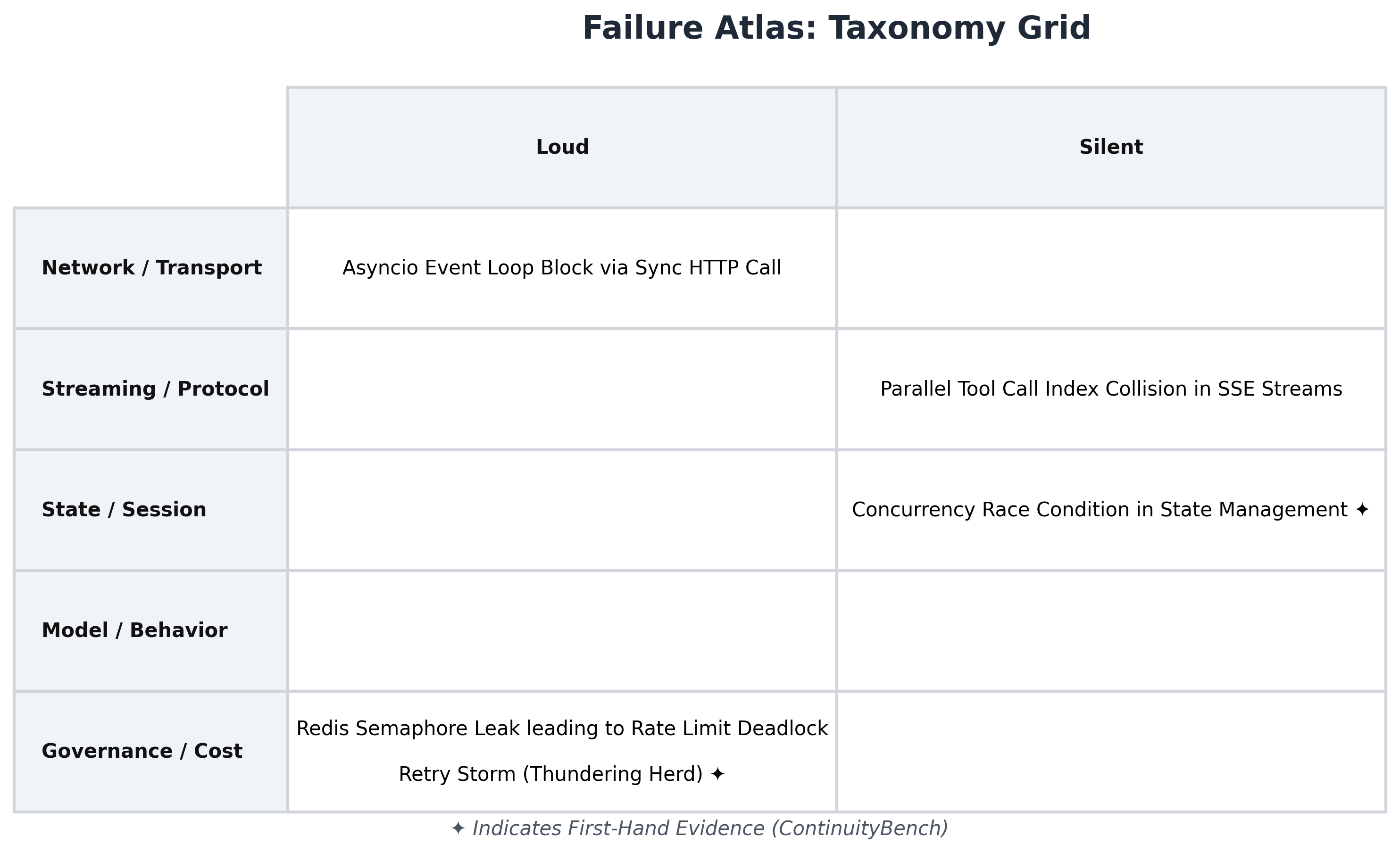}
    \caption{The \fa{} taxonomy grid.  Rows represent the five layers (L1-L5); columns represent Loud vs.\ Silent detectability. Entries marked with \firsthand{} denote first-hand discoveries from \cb{} stress-testing campaigns; unmarked entries are sourced from verified public bug reports.  The Model~Behavior row (L4) is intentionally empty reflecting the difficulty of documenting this failure class to evidence grade.}
    \label{fig:grid}
\end{figure}

\Cref{fig:grid} presents the full $5 \times 2$ taxonomy matrix, populated with the verified entries from our failure catalog.  Each entry is placed in exactly one cell based on its layer of origin and its detectability.

In this section, we present three operationally significant failure modes sourced from our public issue-tracker survey (see \Cref{sec:methodology-sourcing}), selected to illustrate how different layers fail in the wild. The two first-hand case studies discovered during our stress testing are analysed separately in depth in \Cref{sec:case-studies}.

\subsection{Notable Surveyed Failure Modes}
\label{sec:notable-failures}
We contrast one \emph{Silent} protocol failure, which evades standard infrastructure monitoring by corrupting the semantic payload, with two \emph{Loud} failures that trigger hard errors but cause permanent deadlocks if not mitigated.

\subsubsection{Parallel Tool-Call Index Collision in SSE Streams}
\label{sec:ollama}

\begin{quote}
\textbf{Layer:} Streaming/Protocol \quad
\textbf{Detectability:} \textcolor{silentred}{Silent} \quad \textbf{Source:} \href{https://github.com/BerriAI/litellm/issues/33678}{BerriAI/litellm\#33678}
\end{quote}

\paragraph{Mechanism:} When streaming Server-Sent Events (SSE), OpenAI-compatible clients reconstruct parallel tool-call arguments by accumulating text chunks keyed to a specific \texttt{index}. The upstream provider (Ollama) correctly streams one tool call per chunk with the proper index deeply nested. However, proxy layers attempting to derive their own top-level index may inadvertently reset their internal counter per chunk.  Consequently, every tool-call chunk emitted by the proxy is incorrectly stamped with
\texttt{index=0}.

When a model generates multiple parallel tool calls in a single turn, the downstream client receives multiple \texttt{index=0} events. The client obediently concatenates the argument strings of independent tools together (e.g.,\ \texttt{\{"path": "a.rs"\}\{"path": "b.rs"\}}). This corrupted string is silently saved into the client's conversation history. The failure surfaces only on the \emph{next} conversation turn, when the client sends the corrupted history back, producing a \texttt{JSONDecodeError} that misdirects debugging toward the network layer.

\paragraph{Mitigation:} The proxy must maintain a stateful counter across chunks within a single streaming response. The fix was merged in LiteLLM after the issue was reported.

\subsubsection{Redis Semaphore Leak L eading to Rate-Limit Deadlock}
\label{sec:redis}
\begin{quote} 
\textbf{Layer:} Governance/Cost \quad \textbf{Detectability:} \textcolor{loudorange}{Loud} \quad \textbf{Source:} \href{https://github.com/BerriAI/litellm/issues/20256}{BerriAI/litellm\#20256}
\end{quote}

\paragraph{Mechanism:} Gateways often use a Redis-backed semaphore to track in-flight requests for rate limiting (\texttt{max\_parallel\_requests}).  When an incoming request starts, the semaphore increments.  However, if the upstream LLM provider fails with a timeout or 5xx error, and the error-handling path bypasses the decrement logic, the `active' count in Redis is stranded.  As these upstream errors accumulate over time, the semaphore artificially reaches the limit.  At this point, the gateway permanently rejects all new requests for that API key with HTTP~429 errors, even though no requests are actually in flight.

\paragraph{Mitigation:} Enforce a Time-To-Live (TTL) on Redis in-flight counters, or wrap the request execution logic in a strict \texttt{try/finally} block to guarantee decrement on all failure paths.

\subsubsection{Synchronous Health-Check Blocking the Async Event Loop}
\label{sec:healthcheck}

\begin{quote}
\textbf{Layer:} Network/Transport \quad
\textbf{Detectability:} \textcolor{loudorange}{Loud} \quad
\textbf{Source:} \href{https://github.com/BerriAI/litellm/issues/21033}{BerriAI/litellm\#21033}
\end{quote}

\paragraph{Mechanism:} LLM gateways are typically implemented using asynchronous Python (\texttt{asyncio}) to handle high concurrency.  However, some integrations (like database connections or Redis clients) might fall back to synchronous I/O.  If a Kubernetes liveness probe hits a \texttt{/health} endpoint that executes a synchronous database query, that single query blocks the entire main thread's event loop. During this blocked window, no other concurrent requests can progress, leading to cascading timeouts across the application. Because health checks run frequently (e.g.,\ every 10 seconds), this creates persistent micro-outages.

\paragraph{Mitigation:} Ensure all I/O bound operations in the critical path (especially frequent probes) use strictly asynchronous drivers, or offload synchronous calls to a separate thread pool using \texttt{run\_in\_executor}.

\subsection{Structural Observations and Empty Cells}
\label{sec:structural-observations}
Viewing the populated taxonomy matrix as a whole reveals several structural insights about the current state of LLM infrastructure:

\begin{enumerate}
    \item \textbf{No layer is exclusively Loud or exclusively Silent:} The detectability of a failure is not determined by its layer of origin. It is an independent property that depends entirely on whether the failure manifests as an infrastructure signal (an error code, a crash) or a semantic signal (a corrupted context, a malformed JSON payload).  Traditional distributed-systems failures (like the Redis semaphore leak) manifest loudly; LLM-specific protocol failures (like the streaming index collision) manifest silently.
    \item \textbf{The Silent column is sparser but more dangerous:} Only two of our five entries are Silent, yet both are operationally severe. Loud failures get fixed because monitoring catches them; Silent failures persist because nothing flags them. They accumulate as invisible technical debt, systematically under-represented in issue trackers because the infrastructure for noticing them does not yet exist in most deployments.
    \item \textbf{The Model Behavior row is entirely empty:} This is the most conspicuous structural feature of the matrix. Despite widespread anecdotal reports of ``model degradation'' or ``instruction-following drift,'' we were unable to find a single evidence-grade, reproducible bug report that met our inclusion criteria for an infrastructure-level failure in model behaviour. This empty row is a legitimate finding, not a gap in our survey. It highlights that Model Behavior failures are systematically harder to document: they cannot be attributed to a specific code path, their symptoms are diffuse (requiring large-scale statistical evaluation to distinguish from stochastic variation), and detecting them requires continuous semantic evaluation tooling that most organisations lack. The gaps in a taxonomy often point exactly to where the field's observability tooling is weakest.
\end{enumerate}

\section{Case Studies: Fully Characterised Failure Modes}
\label{sec:case-studies}

While the surveyed entries in \Cref{sec:catalog} provide a breadth of in-the-wild examples, they inherently lack the depth of fully controlled experimental observation.  In this section, we present two failure modes discovered first-hand during the development of \cb{} \cite{pandey2026continuitybenchbenchmarksystemsstudy}, an evaluation harness designed to test semantic continuity during provider failover. Because we controlled both the client agents and the mock proxy, we were able to isolate the exact mechanisms of failure, reproduce them under simulated load, and verify their fixes.  We present them here using the \fa{} taxonomy framing to demonstrate how Silent and Loud failures manifest in stateful, multi-provider environments.
\subsection{State Corruption via Concurrency Race Conditions}

\label{sec:race-condition}
\begin{quote}
\textbf{Layer:} State/Session \quad
\textbf{Detectability:} \textcolor{silentred}{Silent}
\end{quote}

The most insidious failure mode discovered during our stress testing was a silent corruption of the conversational state.  It is a classic example of a State/Session layer failure that evades standard infrastructure monitoring.

\paragraph{The Vulnerability:} During early prototype testing at low concurrency ($C=5$ concurrent agent sessions), our failover proxy achieved a near-perfect Continuity Preservation Rate (CPR). However, when subjected to production-scale concurrency ($C=100$), the CPR abruptly collapsed to $\sim$28\%. Analysis of the proxy logs revealed that the fallback provider was returning contextually incorrect responses because the \texttt{messages[]} array it received contained fragments of \emph{other} concurrent conversations. Crucially, the proxy never crashed.  It returned HTTP~200~OK for every request.  The failure was entirely Silent, observable only through the collapse in the semantic continuity metric.

\paragraph{The Mechanism:} The failure stemmed from a classic race condition in the asynchronous evaluation harness, exacerbated by the stateless HTTP abstraction. To minimize network overhead, the proxy maintained a shared, global cache of conversation histories, updating it and transmitting it to the fallback provider upon failover. Under high concurrency, \texttt{asyncio} task interleaving caused parallel requests to mutate the shared history array simultaneously. Because the read-modify-write cycle included an asynchronous yield (awaiting the network call to the LLM provider), the state was not locked. A failover payload destined for Conversation~A would frequently capture the factual anchor established milliseconds earlier by Conversation~B. The coroutine that finished its network call \emph{last} would silently overwrite the shared state, causing an entire turn from the faster coroutine to vanish from the history.

\paragraph{The Fix and Implications:} Resolving this required strictly isolating conversation state by enforcing deep-copies of the \texttt{messages[]} array per-conversation at the local thread level before any asynchronous yielding occurred. While this specific instantiation occurred in our evaluation harness, the architectural vulnerability extends to any stateful proxy design: a gateway that attempts to cache conversation histories internally to avoid client payload bloat must implement robust, per-session read/write locking. Failure to do so results in silent context bleeding across tenant boundaries during failover events.  No standard APM metric (latency, error rate, saturation) will flag this bug.
\subsection{The Failover Retry Storm (Thundering Herd)}

\label{sec:retry-storm}
\begin{quote}
\textbf{Layer:} Governance/Cost \quad
\textbf{Detectability:} \textcolor{loudorange}{Loud}
\end{quote}

In contrast to the silent state corruption, the second failure mode we characterised was overwhelmingly Loud.  It is a textbook example of a Governance/Cost layer failure triggered by upstream instability but amplified by naive downstream retry policies.

\paragraph{The Vulnerability:} During our evaluation of a secondary model as a fallback provider, the entire proxy infrastructure suffered cascading \texttt{ConnectionRefusedError} crashes, permanently wedging the system.  The failures were not distributed randomly; they arrived in massive, synchronous waves that overwhelmed the proxy's connection pool and triggered hard rate limits at the provider level (HTTP~429).

\paragraph{The Mechanism:} When a foundation-model provider experiences a transient failure (e.g.,\HTTP~502 Bad~Gateway), naive HTTP clients within an agentic loop immediately retry the request. In our stress tests, 100 concurrent agent instances were running in parallel.  When the primary provider simulated an outage, all 100 agents encountered the transient failure simultaneously. Because the default retry logic used a fixed interval (e.g.,\ 1 second), all 100 agents scheduled their retries at the exact same moment. This \emph{thundering herd} synchronisation persistently saturated the provider's rate limit on every retry window, rejecting the majority of requests and preventing the provider's load balancer from recovering. Furthermore, because the provider bills for partial completions (tokens generated before a mid-stream failure), each retry wave generated billable work even though the application discarded the incomplete response. This created runaway API spend for zero successful task completions.

\paragraph{The Fix and Implications:} The immediate mitigation is to implement exponential backoff with \emph{jitter} (randomised delay offsets) on all LLM API calls to de-synchronise the retry waves, spreading the load and allowing the provider to recover. Additionally, a global circuit breaker must be enforced at the gateway layer to fail fast and shed load when a provider is degraded, rather than allowing thousands of doomed requests to queue. This failure highlights the critical role of the Governance layer in multi-provider deployments. Failover mechanisms are designed to increase reliability, but without robust backoff and circuit-breaking, they simply convert a transient upstream outage into a self-inflicted Denial of Service (DoS) attack.

\section{Discussion: Why Silent Failures Dominate}
\label{sec:discussion}
The \fa{} catalog is deliberately small, but its composition points to a structural reality about modern LLM-powered applications: the most dangerous failures are not the ones that take the system offline, but the ones that quietly corrupt its internal state while reporting perfect health. The two Silent entries in our catalog the concurrency race condition (\Cref{sec:race-condition}) and the streaming index collision (\Cref{sec:ollama}) exemplify this pattern. We argue that this new class of systems is unusually prone to silent failures due to three architectural mismatches between how these systems are built and how they are monitored.

\subsection{Architectural Mismatches}

\paragraph{Stateless-by-design architecture vs.\ stateful workloads:} Web infrastructure is overwhelmingly designed to be stateless. Load balancers, HTTP proxies, and REST APIs are built on the assumption that Request~A has no bearing on Request~B. Multi-turn LLM applications, however, are deeply stateful; the conversation history is the primary context that dictates model behaviour. When gateways attempt to manage this state internally (as in our race condition case study) or reconstruct it on the fly (as in the SSE index collision), they are fighting the underlying stateless primitives. When these state-management workarounds fail, they do not crash; they simply mutate the payload incorrectly.

\paragraph{Success metrics defined at the transport layer:} Current observability tooling (e.g.,\ Prometheus, Datadog) defines ``success'' at the HTTP level. If a request returns HTTP~200~OK and completes within an acceptable latency budget, the monitoring system considers it a success. But LLM interactions have a semantic payload. If a proxy drops a critical user turn from the context window but returns a perfectly well-formed JSON response containing a hallucinated answer, the HTTP layer reports 100\% success. The success metric is fundamentally decoupled from task correctness.

\paragraph{Lack of established monitoring conventions:} While standard software engineering has decades of conventions for monitoring distributed systems, there is no established convention for monitoring the \emph{semantic integrity} of an LLM pipeline. Developers rely on manual testing or vibes-based evaluation.

\subsection{Continuity as a Motivating Example}
This systemic blindness to semantic failure was the primary motivation behind \cb{}. In our race condition case study (\Cref{sec:race-condition}), the only way we discovered the state corruption was because we were continuously evaluating the \emph{continuity} of the agent's persona. Without a semantic evaluation layer running alongside the infrastructure, the context bleeding across tenant boundaries would have remained invisible. The industry urgently needs observability tools that treat the semantic payload as a first-class citizen, capable of raising alarms when the history is truncated or the model drifts, just as reliably as PagerDuty triggers on a 502 Bad Gateway.

\section{Limitations}
\label{sec:limitations}

\fa{} provides a foundational vocabulary for discussing LLM infrastructure failures, but this initial study has several limitations:

\paragraph{Non-exhaustive catalog:} The catalog currently contains only five entries.  It is by no means an exhaustive survey of all possible failure modes in multi-provider setups. The empty Model~Behavior row, in particular, highlights a gap where anecdotal reports are common but rigorous documentation is scarce.

\paragraph{Varying evidence quality:} While all entries met our strict inclusion criteria, the depth of available evidence varies.  The two first-hand case studies include controlled experimental data and custom reproduction harnesses.  In contrast, the surveyed entries rely on public issue trackers; some are extensively documented by multiple users (e.g.,\ the Redis semaphore leak), while others originate from a single, detailed bug report.

\paragraph{Taxonomy design choices:} The two axes of our taxonomy (Layer and Detectability) are a pragmatic design choice aimed at operational utility.  They do not represent a mathematically provable or mutually exclusive decomposition of all failure space. Other practitioners might reasonably argue for a third axis (e.g.,\ recovery time) or draw the layer boundaries differently.

\paragraph{Self-sourced bias:} Two of the five catalog entries and arguably the most thoroughly characterised ones, were discovered first-hand by the authors during \cb{} evaluations.  This introduces a self-sourcing bias. Our evaluation harness was specifically designed to stress-test state management and retry policies, which naturally surfaced the Race Condition and Retry Storm failures.  Future iterations of the catalog must lean more heavily on community-sourced discoveries to ensure balanced coverage across all layers.

\section{Conclusion}
\label{sec:conclusion}
This paper introduced \fa{}, a structured taxonomy and evidence-grounded catalog of failure modes in multi-provider LLM serving infrastructure. By dissecting failures across five architectural layers and partitioning them by detectability, we demonstrated a critical operational asymmetry: the most dangerous failures in modern agentic systems are overwhelmingly \emph{silent}, returning HTTP~200~OK while corrupting the semantic payload in ways that standard observability tooling cannot detect. We substantiated this taxonomy with deep, reproducible case studies, including the discovery of silent state-corruption race conditions and cascading retry storms that plague naive failover implementations. However, a taxonomy is only as useful as the catalog that populates it. The rapid evolution of LLM proxy layers, stateful tool-calling protocols, and cross-provider routing guarantees that new failure modes will emerge faster than any single research team can document them.  For this reason, the \fa{} catalog is designed as an open, extensible repository. Much like CVE databases or community-curated ``awesome lists,'' \fa{} benefits enormously from collective intelligence.  We invite practitioners, proxy maintainers, and researchers to submit new, evidence-grade catalog entries via pull requests.  By aggregating these hard-won lessons into a shared vocabulary, we can begin building the semantic observability tools necessary to make LLM infrastructure as robust as the web infrastructure that preceded it.

\appendix
\section{Full Failure Catalog}
\label{app:full-catalog}
This appendix provides the structured YAML definitions for all five verified
failure modes included in the initial release of \fa{}.  These entries
serve as the foundational dataset for the taxonomy presented in
\Cref{sec:taxonomy}.  The machine-readable YAML files, along with their
corresponding reproduction scripts and setup instructions, are available in
the open-source companion repository.
We encourage practitioners and researchers to submit pull requests adding
new, evidence-grade failure modes to this catalog using the schema
demonstrated below.
\subsection{Entry 1: Concurrency Race Condition}
\begin{lstlisting}[language=bash]
id: concurrency-race-condition
name: Concurrency Race Condition in State Management
layer: state-session
detectability: silent
summary: Asynchronous turn processing overwrites or drops previous turns during concurrent requests, leading to silent continuity loss.
mechanism: When multiple agents share a conversation state store (like a Python dictionary) without proper read-modify-write locking, async task interleaving causes concurrent turns to silently overwrite each other upon returning from the LLM network call.
evidence:
  - type: first-hand
    source: ContinuityBench Phase 2 ($C=100$)
    description: 72% drop in Continuity Preservation Rate under high concurrency due to silent turn loss.
reproduction:
  available: true
  path: reproductions/race-condition/repro.py
  notes: Runs a minimal asyncio script showing shared vs isolated state mutation.
mitigation: Wrap state mutations in asyncio.Lock or use deep-copy isolation per tenant session.
\end{lstlisting}
\subsection{Entry 2: Parallel Tool-Call Index Collision}
\begin{lstlisting}[language=bash]
id: ollama-streaming-tool-index-zero
name: Parallel Tool-Call Index Collision in SSE Streams
layer: streaming-protocol
detectability: silent
summary: Proxy resets chunk index counter, causing multiple parallel tool calls to merge into a corrupted JSON string.
mechanism: Ollama correctly streams multiple tool calls per turn using nested indices. The proxy layer's SSE iterator incorrectly derives its own index from chunk-local state, resetting to index=0 for every chunk. The client concatenates all arguments into a single malformed payload.
evidence:
  - type: github-issue
    source: https://github.com/BerriAI/litellm/issues/33678
    description: Issue detailing the chunk iterator logic flaw and JSONDecodeError symptom.
reproduction:
  available: true
  path: reproductions/ollama-index-bug/repro.py
  notes: Feeds captured Ollama chunks through the buggy iterator locally.
mitigation: Preserve the provider's native index or maintain stateful cross-chunk counters in the proxy.
\end{lstlisting}
\subsection{Entry 3: Failover Retry Storm (Thundering Herd)}

\begin{lstlisting}[language=bash]
id: retry-storm-thundering-herd
name: Failover Retry Storm (Thundering Herd)
layer: governance-cost
detectability: loud
summary: Uncoordinated, fixed-interval retries across multiple agents during a provider outage trigger a self-inflicted denial of service.
mechanism: When a provider throws a transient 5xx error, concurrent agents all schedule a retry at a fixed interval (e.g. 1s). The synchronized retry wave saturates the provider's rate limit repeatedly, preventing recovery and generating runaway billable API spend for partial completions.
evidence:
  - type: first-hand
    source: ContinuityBench failover stress tests
    description: Cascading ConnectionRefusedErrors in the proxy during Gemini fallback evaluation.
reproduction:
  available: true
  path: reproductions/retry-storm/repro.py
  notes: Simulates 100 concurrent failovers against a 15 req/min rate-limited provider.
mitigation: Implement exponential backoff with jitter on all LLM API calls and global circuit breaking.
\end{lstlisting}

\subsection{Entry 4: Redis Semaphore Leak}
\begin{lstlisting}[language=bash]
id: litellm-max-parallel-deadlock
name: Redis Semaphore Leak Leading to Rate-Limit Deadlock
layer: governance-cost
detectability: loud
summary: Upstream errors bypass semaphore decrement logic, stranding active counts and permanently locking the gateway.
mechanism: The gateway uses Redis to track active in-flight requests. If an upstream provider times out or errors, and the error path misses the decrement step, the count permanently leaks. Once the leaked count hits the configured max_parallel_requests limit, all new requests are rejected with 429s.
evidence:
  - type: github-issue
    source: https://github.com/BerriAI/litellm/issues/20256
    description: User reported permanent 429 lockouts resolved only by manual Redis flush.
reproduction:
  available: false
  path: reproductions/README.md
  notes: Requires a running Redis instance and specific error-injection proxy setup. mitigation: Enforce Redis TTLs on active request counters or wrap execution in strict try/finally blocks.
\end{lstlisting}
\subsection{Entry 5: Asyncio Event Loop Block}
\begin{lstlisting}[language=bash]
id: litellm-sync-base64-healthcheck-hang
name: Asyncio Event Loop Block via Synchronous HTTP Call
layer: network-transport
detectability: loud
summary: A synchronous HTTP call in the critical path blocks the async event loop, causing cascading timeouts and health-check failures.
mechanism: A function converting image URLs to base64 uses a synchronous HTTP client. When an unroutable IP is provided, the call hangs for the 2-minute socket timeout. This blocks the main Python asyncio event loop, preventing all other concurrent requests and liveness probes from executing.
evidence:
  - type: github-issue
    source: https://github.com/BerriAI/litellm/issues/24788
    description: Issue detailing Kubernetes probe failures and cascading pod restarts.
reproduction:
  available: false
  path: reproductions/README.md
  notes: Requires a Kubernetes environment with active liveness probes to demonstrate the crash loop. mitigation: Enforce short connect timeouts and offload synchronous I/O to a thread pool via run_in_executor.
\end{lstlisting}

\nocite{*}
\bibliographystyle{unsrt}  
\bibliography{references}

\end{document}